# Content-Based Image Retrieval Using Multiresolution Analysis Of Shape-Based Classified Images

I. M. El-Henawy[1], Kareem Ahmed[2]

Computer Science Department, Faculty of Computer Science, ZAGAZIG university,

Henawy2000@yahoo.com

Multimedia Department, Faculty of Computer Science, BSU University,

Kareem_ahmed@hotmail.co.uk

## Abstract

Content-Based Image Retrieval (CBIR) systems have been widely used for a wide range of applications such as Art collections, Crime prevention and Intellectual property. In this paper, a novel CBIR system, which utilizes visual contents (color, texture and shape) of an image to retrieve images, is proposed. The proposed system builds three feature vectors and stores them into MySQL database. The first feature vector uses descriptive statistics to describe the distribution of data in each channel of RGB channels of the image. The second feature vector describes the texture using eigenvalues of the 39 sub-bands that are generated after applying four levels 2D DWT in each channel (red, green and blue channels) of the image. These wavelets sub-bands perfectly describes the horizontal, vertical and diagonal edges that exist in the multi-resolution analysis of the image. The third feature vector describes the basic shapes that exist in the skeletonization version of the black and white representation of the image. Experimental results on a private MYSQL database that consists of 10000 images; using color, texture, shape and stored relevance feedbacks; showed 96.4% average correct retrieval rate in an efficient recovery time.

## Indexing terms/Keywords

CBIR; wavelet; Sobel; Color; texture; Fourier; Shape; bp-FFN.

## INTRODUCTION

The rapid development in the field of information technology and the appearance of a huge number of specialized devices for capturing and storing images, has led to the emergence of the so-called multimedia databases. A multimedia database contains various data types such as text, images, animation sequences, audio and video. Unfortunately, multimedia data are not stored in a standard format, which makes indexing and content-based retrieval impossible. MySQL, which is the world's second most widely used open-source relational database management system (RDBMS), offers to store multimedia objects using BLOB (binary large object) data type. BLOB, which is a group of binary data stored as a single entity in a DBMS, is used to store a variable amount of binary data without any predefined character data set. There are four types of BLOB data according to the maximum size of the stored data: TINYBLOB, BLOB, MEDIUMBLOB and LONGBLOB. The BLOB data is used in this research to store images in the MYSQL database and to reduce needs for further file processing and indexing schemas. The main advantages of this approach are: all data are stored into one place, easier and faster storage/retrieval, and eliminating the needs for any additional modules for file searching and file indexing. In order to make the use of multimedia databases more efficient, users need to browse, search and retrieve images from the large databases of digital images. Content-based image retrieval (CBIR), also known as query by image content (QBIC) and content-based visual information retrieval (CBVIR) is the application of computer vision techniques to retrieve digital images from large databases[1]. CBIR systems performs the search by analyzing the contents of the image (such as colors, shapes, textures, or any other information that can be derived from the image itself) rather than using the metadata (keywords or tags) associated with it. The techniques and algorithms that are used in CBIR systems originate from fields such as computer vision, statistics and pattern recognition. Despite that currently many web based image search engines rely purely on metadata, but this approach has a lot of difficulties such as inconsistencies in the retrieved results, as well as the difficulty of obtaining every keyword that describes stored images in the large database. Therefore, the existence of an automated system that is able to filter images based on their content would provide better indexing and more accurate retrieval. CBIR systems are useful for a wide range of applications including: art collections, crime prevention and intellectual property, to name only a few. Another CBIR significant application is in the medical field[8, 14], due to the huge amount of digital images that are generated every day. Some software producers are aggressively trying to integrate CBIR techniques into the filtering and law enforcement markets to identify and censor images with skin-tones and shapes that could indicate the presence of nudity[4]. Different CBIR systems may accept different types of user queries, such as

- The user may supply an existing image.
- The user may select an image from a random list.
- The user may use a digital pen or stylus to draw a rough approximation of the image he/she is looking for.

- The user may navigate through hierarchical categories.
- The user may select a specific region within the image rather than the entire image.
- The user may supply multiple example images
- The user may supply a visual sketch.
- The user may supply directly the specification of image features

Robust CBIR systems should allow users to provide relevance feedback, where users progressively improve search results by marking images in the results as "relevant" or "not relevant" to the search query, then repeating the search after submitting these feedbacks.

## STATE-OF-THE-ART

The term CBIR was used in 1992 by T. Kato [2] to describe automatic retrieval of images from a database, based on the colors and shapes. After that, it has been used to describe the process of retrieving desired images from a large collection on the basis of syntactical image features[3]. There are many hybrid approaches [14,15] that use color, texture and shape-based features for visual information retrieval. There are many publicly available CBIR search engines which examines the contents (pixels) of their images to retrieve results that match a particular query. Some examples of CBIR search engines that use keywords or an external image to retrieve similar images are Google Image Search, Yandex Image Search, Baidu Image Search, akiwi, ALIPR, IKONA, Lucignolo, PIRIA, VIRaL, SHIATSU, MiPai similarity search engine and Visual Recognition Factory. There are also a number of commercial desktop applications such as NoClone 2013 and imense which are used to find similar images which have been resized, clipped, rotated, slightly modified or saved in a different format. M. Lux [18] developed a light weight open source Java library for CBIR called LIRe (Lucene Image Retrieval) which can be integrated in applications without relying on a database server. Some examples of CBIR search engines that use only an external image to retrieve similar images are Visual Image Search, Chic Engine, Anaktisi, BRISC, Caliph & Emir, GNU Image Finding Tool, img(Rummager), imgSeek, IOSB, MUVIS, Pixcavator, pixolu, SIMBA, Windsurf, PIBE, TinEye, IMMENSELAB and Macroglossa Visual Search. There are a lot of CBIR search engines that use only the metadata (keywords) associated with images to retrieve similar images, such as Imense Image Search Portal, QuickLook, TagProp, Imprezzeo Image Search, Incogna Image Search, Like (Shopping & fashion based CBIR engine), Empora, Shopachu, Tiltomo and eBay Image Search. There are also CBIR search engines that allow you to select one or several colors and retrieve images that composed primarily of the selected color(s), such as TinEye lab.

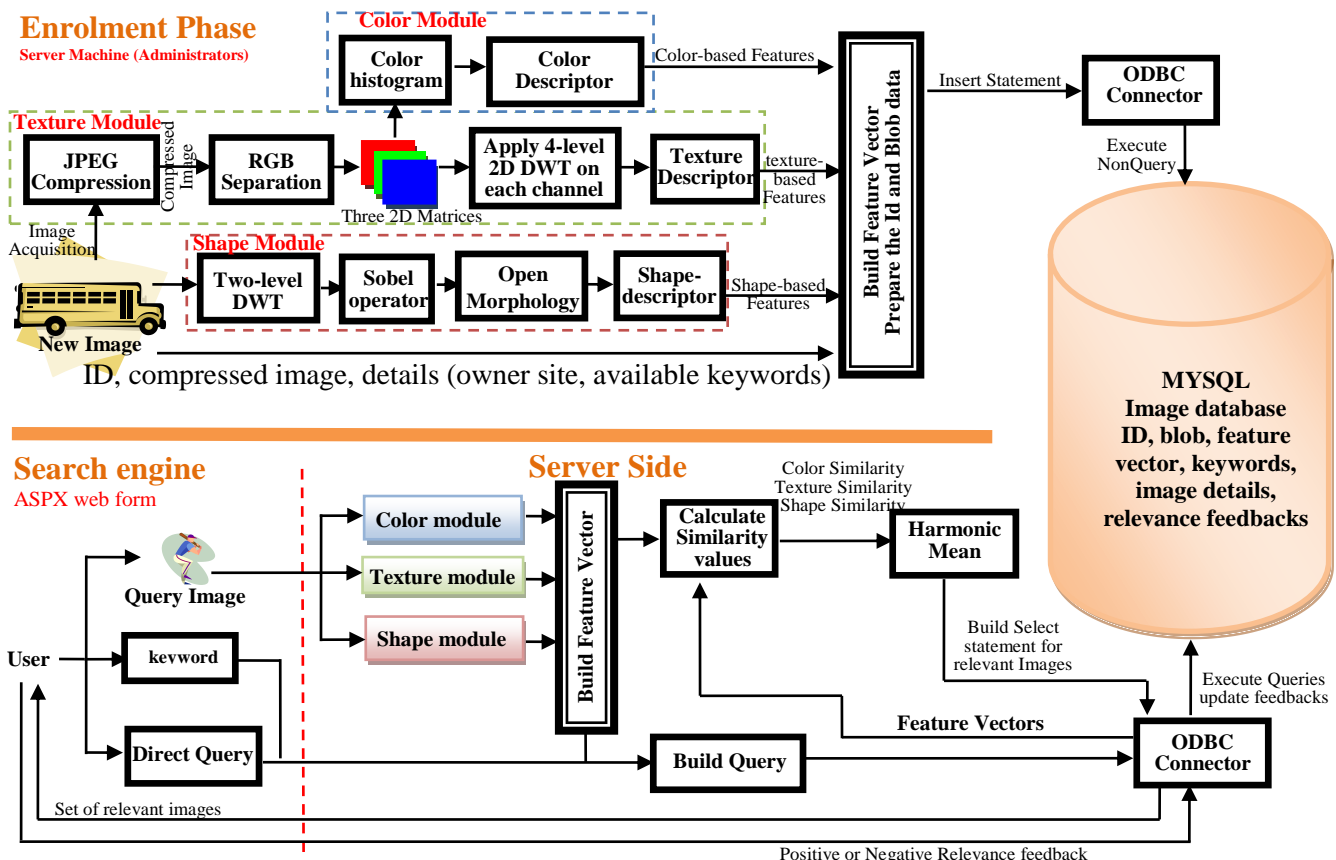

**Fig 1: General framework of the proposed Method**

## GENERAL FRAMEWORK OF THE PROPOSED METHOD

In this paper, a novel Content-Based Image Retrieval system, which utilizes visual contents (color, texture and shape) of an image to retrieve and classify images, is proposed. The proposed system builds three feature vectors from the images before storing them into MySQL database. The first feature vector describes the distribution of each color in the RGB representation of the image. The second feature vector describes the texture as well as the horizontal, vertical and diagonal edges that exist in the multi-resolution analysis of the image. The third feature vector describes the basic shapes that exist in the skeletonization version of the black and white representation of the image.

The retrieval of images based on the colors they contain is often achieved through comparing color histograms of the images, but there are many other techniques to achieve it. The advantage of this approach is that it is scale and rotation invariant. The color histogram identifies the number of pixels within an image holding specific intensity values. Many researchers suggested to segment the image into several regions and calculate the histogram for each region.

Image Texture presents information about visual patterns in images (or in a specific region in the image) and how they are spatially arranged. Textures are considered one of the most important features in CBIR systems. Textures are particularly useful for searching visual databases because they are replications, symmetries and combinations of various basic patterns or local functions, usually with some random variation[6]. In this research, multi-resolution analysis is performed to extract feature vector that depicts the texture in the image.

Shape does not mean the shape of the whole image but it means the shape of a particular object that exists in the image. Shapes are always estimated using segmentation, morphological operations or edge detection masks. The shape is extracted by marking the boundary edges of it. An edge is defined as a jump in intensity. Thus the edges are useful for segmentation, registration, and identification of objects in the image. H.B. Kekre et al. [7] described several edge detection masks for shape extraction like Sobel, Roberts, Prewitt and Canny gradient operators.

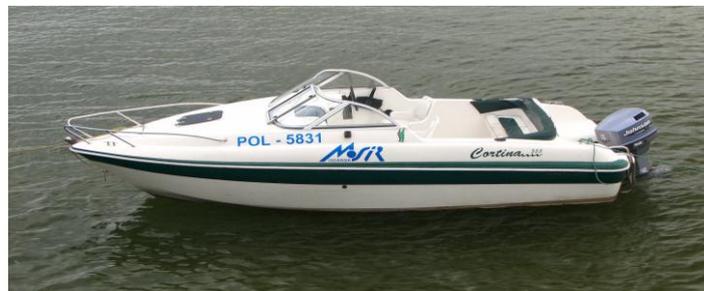

**Fig 2: Original Image (Motor boat in Gdansk-Brzeźno, Poland)**

In this section, the proposed system is introduced and its main parts are explained in details. Figure 1 shows the general framework of the proposed system which will be discussed later in this section.  The system performs two major functions, the first is image registration (or image enrollment) and the second is content-based image retrieval. Image registration assigns a unique number (primary key) for the image, then compresses and stores the JPG image into the BLOB field in the database. It also extracts feature vector that represents the image and stores it in the database to be used later for comparisons and retrieval processes. It also stores any keywords that are available for the image and the code of the basic category of the image. Finally, it stores any additional information that are associated with the image, such as the owner website, copyrights, and any other details such as photography date and place. In the other hand, the content-based image retrieval is performed in two stages. The first stage is done at the client-side through an aspx form, where the user can upload an image which he/she wants to search for similar images. The user can also browse the images stored in the database through typing keywords or manually submitting the feature vector of the image he/she wants to search. The second stage is done at the server-side, where the server extracts the feature vector from the query image and compares it with the feature vectors stored in the database, then determines the degree of similarity between compared vectors. After that, the system retrieves the similar images, ranks them and sends them back to the user's browser. Finally, the user can provide relevance feedback to be used to improve the future image retrievals.  The next subsections describe each module that is used to extract color, texture and shape features.

    **I.  Color-Based  Features Extractor Module**

When adding a new image to the database, the system first performs JPEG compression to reduce the space required to store the image into the blob field in the database. The JPEG is able to achieve 10:1 compression with little or no perceptible loss in image quality. After that, color Separation is performed to extract Red, Green and blue Channels. Figure 2 shows an original image which is Motor boat in Gdansk-Brzeźno, Poland. Figures 3a, 3b and 3c show red, green and blue channels extracted from the original image in figure 2.

(a)

(b)

(c)

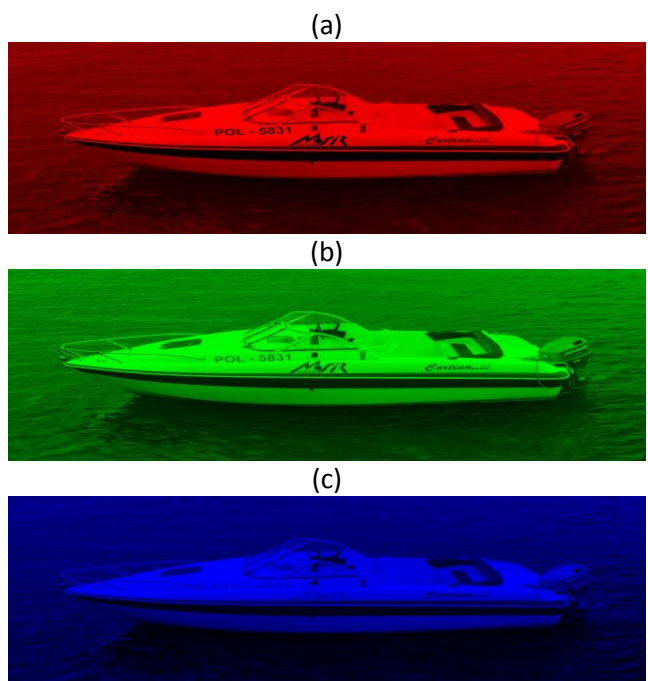

**Fig 3: Red, Green and Blue Channels of the original Image**

The second step is to calculate the color histograms for each channel. Three histogram charts are produced from this process, as shown in figure 4. Each histogram is a table that contains intensity values and the frequency of each intensity value in one channel. Ten features are extracted from each one of the resulting three histogram tables. Thus, a total of 30 (10 features × 3 Red, Green, Blue colors) features will be used to depict the distribution of different colors in the image. The extracted features are: Mean, Median, Mode, First Quartile, Third Quartile, 60th Percentile, Standard Deviation, Interquartile Range (IQR), Range And Skewness. The 60th percentile is the value below which 60 percent of the observed intensity values are occurred. The observed values may be Symmetric (Bell-Shaped), Skewed Left or Skewed Right, so Skewness is used to determine if a distribution has a single long tail.

## II. Texture-based features extractor module

Textures have been widely used by many researchers as one of the most visual primitives that are used to analyze different regions in the image to find homogenous areas. There are many approaches for extracting texture-based features and using them for visual information retrieval. I.J. Sumana et al. [10] present a retrieval performance comparison between generalized Gaussian density (GGD) texture features, curve let texture features, wavelet texture features and wavelet GGD texture features. It was shown that texture analysis is an essential element in visual information retrieval. There are no exact definition for the texture because there are a very wide spectrum of textures. A region in an image is said to have a constant texture if a set of local statistics are constant, slowly varying or approximately periodic. Mann-Jung Hsiao [11] classified the approaches used to describe the texture of a region into two main categories: statistical and structural. The statistical approach is based on spatial frequencies and produces characterizations of textures as smooth, coarse, grainy or any other type. Some examples of statistical approaches are autocorrelation function, gray-level co-occurrence matrix, Fourier texture analysis and edge frequency. The structural approach describes the texture based on arrangement of image primitives such as regularly spaced parallel lines.

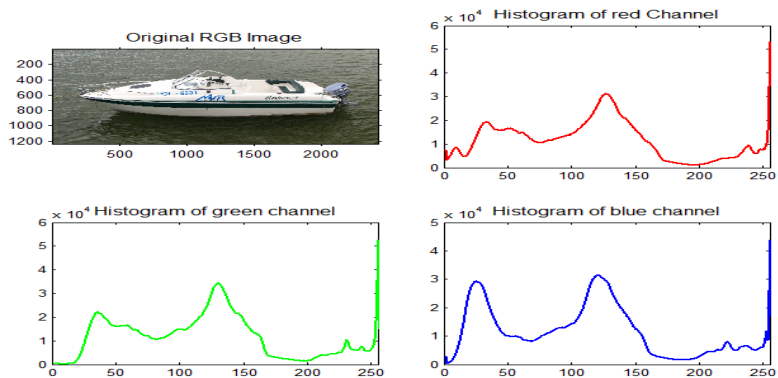

**Fig 4: Histogram charts plots the number of pixels in each channel (vertical axis) with intensity values (horizontal axis).**

It is well known that images contain distinct features at various levels of detail or scales, so small-sized objects are preferable to be analyzed at high resolutions and large-sized objects are preferable to be analyzed at low resolutions. In this research, the texture is studied using multi-resolution analysis, in the wavelet domain. Multi-resolution analysis is concerned with analyzing the image at different frequencies with different resolutions. The basic advantage of multi-resolution analysis is that it provides information simultaneously in both the spatial and frequency domains. Multi-resolution analysis was first designed for one-dimensional time-based signals to give good time resolution and poor frequency resolution at high frequencies and good frequency resolution and poor time resolution at low frequencies[9]. In the beginning, the proposed system converts a not square image into a square image to be able to calculate eigenvalues of resulting DWT coefficient matrices later. This conversion is performed by zero-padding or simply adding zero-rows or zero-columns to end of each matrix (red, green and blue matrices) to increase its size. If the image is of size N×M; where N is the number of rows and M is the number of columns; and N is greater than M, then the system adds zero N-M columns to each channel, otherwise, the system adds zero M-N rows to each channel. The resulting of this conversion is a square image. Then the system analyzes each channel of the three RGB channels using four level two-dimensional Haar Discrete Wavelet Transform (DWT), as shown in figure 5. The advantages of using the 2D Haar are orthogonality, biorthogonalality, symmetry, good spatial and frequency localization, ability to perform multi-resolution decomposition and compactly supported. Figure 6 shows the four sub-bands LL, HL, LH, and HH that are generated after applying four level of DWT in the input channel. The 2D DWT is performed by applying 1D DWT in each row of the channel to produce an intermediate result followed by applying the same 1-D DWT in each column of the intermediate result to produce the final result (wavelet coefficients). The low-pass sub-band LL (also called approximation matrix AP) represents a 4:1 sub-sampled version of the original channel. After applying four level 2D DWT on each channel of the three red, green and blue channels, 39 matrices are produced (3n+1 for each channel where n is the number of levels × 3 channels). Finally, the maximum eigenvalue of each matrix is computed and stored in the texture feature vector, and hence, a 39-tuple feature vector is produced. Eigenvalues [12], also known as characteristic roots, proper values, latent roots or characteristic values, are a set of scalars associated with a linear system of equations. Eigenvalues are very important and have been widely used for many applications such as stability analysis, the physics of rotating bodies, and small oscillations of vibrating systems. The 39-tuple feature vector is then stored with the corresponding BLOB data in the multimedia database to be used later to examine texture features

### III. Shape-based features extractor module

Shape is considered one of the most important and powerful feature that is used for image classification and indexing. A. Banerjee [13] proposed a new method of shape based image retrieval which uses modulus and amplitude parts of Fourier transform to extract features. M.A.Z. Chahooki [16] reported that the fusion of contour and region-based shape description significantly improves retrieval performance. In this paper, shape is selected as a major feature that is used to classify and index images in the database. Firstly, nine categories for images are determined but these categories can be increased in the future. The categories are boats, animals, cartoon, automobiles, human, trees, buildings, computers and trains. A two layer Back-Propagation Feed-Forward Neural Network (BP-FFN) is used to identify the category of a given shape. When a new image is enrolled into the database, the basic shape in the image is extracted and entered to the BP-FFN, which classifies the extracted into one of the nine categories and stores the category code with the image into the database.

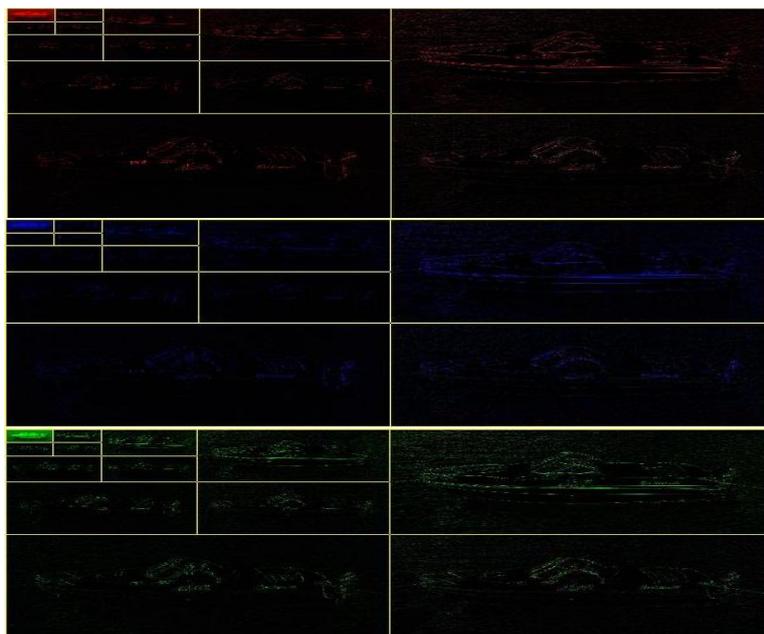

**Fig 5: Four level 2D Haar DWT of each channel of RGB channels.**

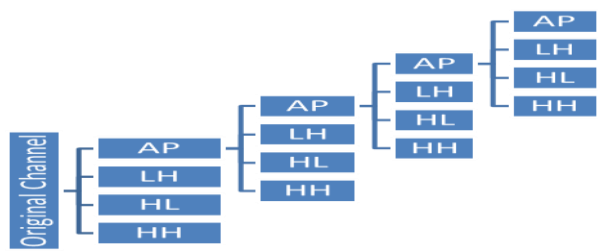

**Fig 6: The four sub-bands that are generated after applying four level of DWT in the input channel.**

Later, when a query image is presented for retrieval of similar images, its basic shape is extracted and recognized by the BP-FFN as belonging to one of the nine predetermined categories. The system is then compares the query image with the images in the database that belong to the same category as the query image. The basic steps of the algorithm proposed on this paper can be summarized as follows:

1. *Two-level 2D Haar Discrete Wavelet Transform.*
2. *Reconstruct the original image using only the approximation matrix at the second level.*
3. *Convert the resulting image to black and white.*
4. *Apply the Sobel operator on the resulting image.*
5. *Use open morphology to depict the skeletonization of the extracted shape.*
6. *Use Fast Fourier transform to make the shape robust against translation, rotation and scaling.*
7. *Shift zero-frequency component to center of spectrum*
8. *Take the logarithm of the absolute values after adding one to each value to avoid logarithm of zero.*
9. *Use the highest 30 coefficients to train or test the BP-FFN.*
10. *Use the output of the BP-FFN to specify the image category.*
11. *Execute the Color-based features extractor module and the Shape-based features extractor module, that were discussed previously in this section, on the query image.*
12. *Calculate the color similarity and texture similarity between the query image and the images stored in the database that belong to the category specified at step 10.*
13. *Use the harmonic mean to merge color and texture similarity into one value.*
14. *If the final score is above a certain threshold then retrieve the image, else try the next image on the same category.*

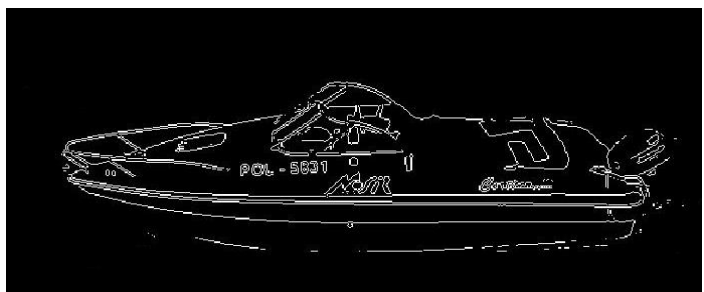

**Fig 7: The basic shape of the image in figure 2.**

Edge detection is the first step that is performed to extract shape-based features. In this paper the Sobel operator is used for edge detection. Sobel operator [17] is a discrete differentiation operator that computes an approximation of the gradient of the image intensities. At each pixel, the Sobel operator produces either the corresponding gradient vector or the norm of this vector. The choice of the Sobel operator is because it is cost-efficient in terms of computations since it depends on convolving the image with a small, separable, and integer valued filter in horizontal and vertical directions. The second step is to apply the open morphology on the intermediate results that were produced using the Sobel operator. Generally, morphology is always used to modify, discover and determine the shapes of objects in the image. Figure 7 shows the results of applying steps from 1 to 5 on the proposed algorithm on the image in figure 2. The basic advantage of applying morphological operations is to represent and model the binary image by parts of its skeleton. The resulting image is a thinned version of the image that contains complete information about its shape and its size. To make the description of the extracted shape robust against translation, rotation, and scale, Fast Fourier Transform is applied to the resulting skeletonization version of the image. Figure 8 displays the resulting coefficients after applying steps from 1 to 8 on the image in figure 2. On the other hand, The BP-FFN is used to describe a specific category for the query image to limit the number of comparisons and matching to a compact number which significantly affects the speed and efficiency of the retrieval system. After applying all steps from 1 to 11, the problem is reduced to just a comparison between two color-based feature vectors and 2 texture-based feature vectors, which is accomplished using a simple Euclidean distance. After that the harmonic mean is used to merge the two similarity values and then the result is compared to a threshold to decide whether to retrieve the current image or not. The next section describes the list of experiments that are performed to show the effectiveness of the proposed method. After each trial for image retrieval, the user is optionally allowed to submit a positive or negative feedback; that is used to update the category code of the retrieved image; in order to improve future retrievals.

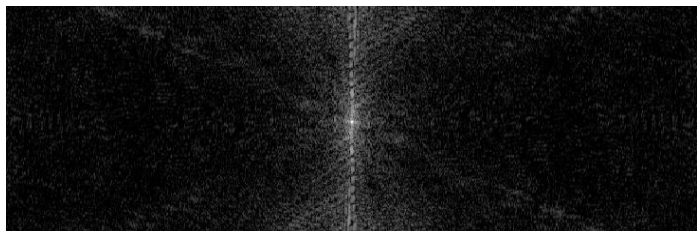

**Fig 8: The logarithm of the absolute FFT coefficients**

## EXPERIMENTAL RESULTS

The methodology that was used to test the efficiency of the proposed system is based on performing ten experiments on images of each category, so 90 (10 image searches × 9 categories) content-based image retrieval experiments were performed. The used database consists of 10000 images, each image is belonging to only one of nine categories which are boats, animals, cartoon, automobiles, human, trees, buildings, computers and trains. Table 1 shows the average Correct Retrieval Rate (CRR) and the average False Retrieval Rate (FRR) of the experiments that are performed on images using color and texture features without predetermining the corresponding image category. Table 2 shows the results of ten experiments performed on each category when including the shape-based features and using them by the BP-FFN to determine the category that will be searched inside. The performance of the system on the cartoon category was decreased because there are no standard shapes for this category which indicates that the system still has some problems with ambiguous categories.

**Table 1. Average CRR and FRR using only Color and Texture Features**

| Exp. ID | Query Image Category | No of trials | Average CRR | Average FRR |
|---|---|---|---|---|
| 1 | boats | 7 | 95.35 | 4.65 |
| 2 | animals | 5 | 88.13 | 11.87 |
| 3 | cartoon | 9 | 90.56 | 9.44 |
| 4 | automobiles | 8 | 95.64 | 4.36 |
| 5 | human | 4 | 85.41 | 14.59 |
| 6 | trees | 5 | 84.34 | 15.66 |
| 7 | buildings | 4 | 87.28 | 12.72 |
| 8 | computers | 2 | 90.78 | 9.22 |
| 9 | trains | 6 | 89.54 | 10.46 |
| | **Average Performance** | | 89.67 | 10.33 |

**Table 2. Average CRR and FRR using Color, Texture and shape Features**

| Exp. ID | Query Image Category | No of trials | Average CRR | Average FRR |
|---|---|---|---|---|
| 1 | boats | 10 | 98.71 | 1.29 |
| 2 | animals | 10 | 92.43 | 7.57 |
| 3 | cartoon | 10 | 85.77 | 14.23 |
| 4 | automobiles | 10 | 97.44 | 2.56 |
| 5 | human | 10 | 94.32 | 5.68 |
| 6 | trees | 10 | 94.55 | 5.45 |
| 7 | buildings | 10 | 92.77 | 7.23 |
| 8 | computers | 10 | 95.36 | 4.64 |
| 9 | trains | 10 | 98.57 | 1.43 |
| | **Average Performance** | | 94.44 | 5.56 |

The classification of the images were performed automatically by the system after training the neural network on different shapes of each category. Most of the consumed time was for manually selecting appropriate images that contains several shapes to describe a specific category, in order to train the BP-FFN with most of the available shapes. After that an automatic server-side desktop application was developed to automatically extract shape-based features and use it to identify the most suitable category for an image (category_code) and then extracts color-based and texture-based features for the current image and finally stores the image_ID, Blob_image_data, category_code, color-based feature vector and texture-based feature vector into the database. The program is organized to perform all these operation automatically on each image that is existed on a specific folder. Finally, the average CRR on all experiments reached 96.4% when including the positive or negative relevance feedback that are submitted by the user after each trial.